\documentclass[10pt,twocolumn,letterpaper]{article}

\usepackage{cvpr}              

\usepackage[accsupp]{axessibility}  
\usepackage{graphicx}
\usepackage{amsmath}
\usepackage{amssymb}
\usepackage{booktabs}
\usepackage{multirow}
\usepackage{url}            
\usepackage{amsfonts}       
\usepackage{nicefrac}       
\usepackage{microtype}      
\usepackage{xcolor}         
\usepackage{colortbl}

\usepackage{makecell} 

\usepackage[pagebackref,breaklinks,colorlinks]{hyperref}

\usepackage[capitalize]{cleveref}
\crefname{section}{Sec.}{Secs.}
\Crefname{section}{Section}{Sections}
\Crefname{table}{Table}{Tables}
\crefname{table}{Tab.}{Tabs.}
\definecolor{mygray}{gray}{.94}

\begin{document}

\title{DSVT: Dynamic Sparse Voxel Transformer with Rotated Sets}

\author{
Haiyang Wang$^{1}$\footnotemark[1] ~~~~Chen Shi$^{5}$\footnotemark[1] ~~~~Shaoshuai Shi$^{2}$\footnotemark[2] ~~~~Meng Lei$^{1,4}$  \\
~~~~Sen Wang$^{3}$  ~~~~Di He$^{5}$ ~~~~Bernt Schiele$^{2}$  ~~~~Liwei Wang$^{1,5}$\footnotemark[2]   \\
{\normalsize
{$^1$}Center for Data Science, Peking University }\\
{\normalsize
{$^2$}Max Planck Institute for Informatics, Saarland Informatics Campus ~~ {$^3$}Huawei ~~{$^4$}Zhejiang Lab}\\
{\normalsize{\hspace*{-18pt}}
}
{\normalsize~~{$^5$}National Key Laboratory of General Artificial Intelligence, School of Intelligence Science and Technology, Peking University}\\
{\tt\small \{wanghaiyang@stu, shichen@stu, leimeng@stu, dihe@, wanglw@cis\}.pku.edu.cn}\\
{\tt\small \{sshi, schiele\}@mpi-inf.mpg.de ~~wangsen31@huawei.com}
}

\maketitle
\renewcommand{\thefootnote}{\fnsymbol{footnote}}
\footnotetext[1]{Equal contribution.}
\footnotetext[2]{Corresponding author.}

\begin{abstract}
Designing an efficient yet deployment-friendly 3D backbone to handle sparse point clouds is a fundamental problem in 3D perception. Compared with the customized sparse convolution, the attention mechanism in Transformers is more appropriate for flexibly modeling long-range relationships and is easier to be deployed in real-world applications. However, due to the sparse characteristics of point clouds, it is non-trivial to apply a standard transformer on sparse points.  In this paper, we present Dynamic Sparse Voxel Transformer (DSVT), a single-stride window-based voxel Transformer backbone for outdoor 3D perception. In order to efficiently process sparse points in parallel, we propose Dynamic Sparse Window Attention, which partitions a series of local regions in each window according to its sparsity and then computes the features of all regions in a fully parallel manner. To allow the cross-set connection, we design a rotated set partitioning strategy that alternates between two partitioning configurations in consecutive self-attention layers. To support effective downsampling and better encode geometric information, we also propose an attention-style 3D pooling module on sparse points, which is powerful and deployment-friendly without utilizing any customized CUDA operations. Our model achieves state-of-the-art performance with a broad range of 3D perception tasks. More importantly, DSVT can be easily deployed by TensorRT with real-time inference speed (27Hz). Code will be available at \url{https://github.com/Haiyang-W/DSVT}.
\end{abstract}

\begin{figure}[t]
\vspace{-6pt}
  \centering
   \includegraphics[width=0.98\linewidth]{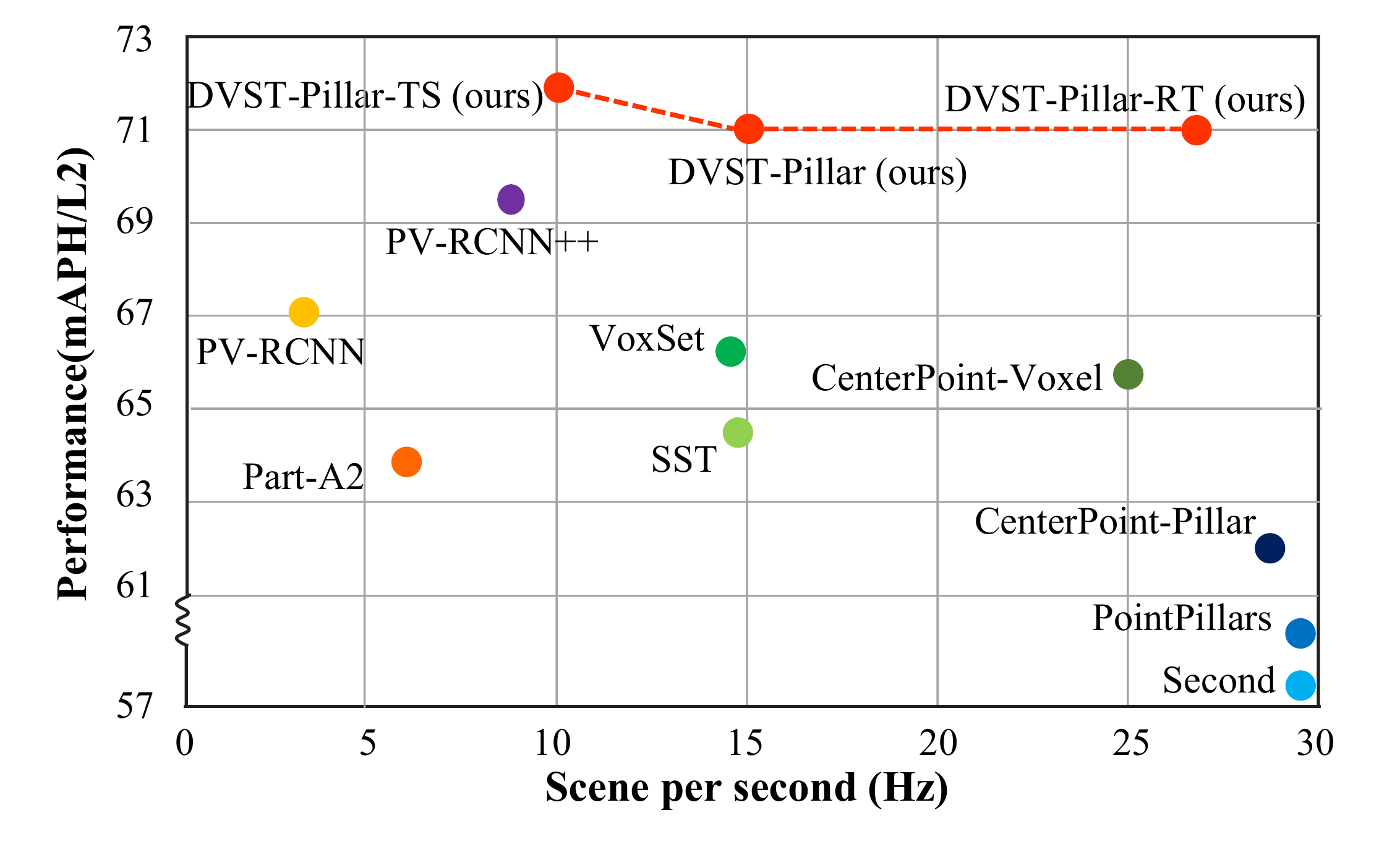}
    \vspace{-14pt}
    \caption{Detection performance (mAPH/L2) vs speed (Hz) of different methods on Waymo~\cite{PeiSun2020ScalabilityIP} validation set. All the speeds are evaluated on an NVIDIA A100 GPU with AMD EPYC 7513 CPU. }
   \label{fig:speeddemo}
   \vspace{-16pt}
\end{figure}

\vspace{-10pt}
\section{Introduction} \label{sec:intro}
3D perception is a crucial challenge in computer vision, garnering increased attention thanks to its potential applications in various fields such as autonomous driving systems~\cite{bansal2018chauffeurnet, wang2019monocular} and modern robotics~\cite{zhu2017target, wang2021collaborative}. In this paper, we propose DSVT, a general-purpose and deployment-friendly Transformer-only 3D backbone that can be easily applied to various 3D perception tasks for point clouds processing.

Unlike the well-studied 2D community where the input image is in a densely packed array, 3D point clouds are sparsely and irregularly distributed in continuous space due to the nature of 3D sensors, which makes it challenging to directly apply techniques used for traditional regular data. To support sparse feature learning from raw point clouds, previous methods mainly apply customized sparse operations, such as PointNet++~\cite{qi2017pointnet,qi2018pointnnetplus} and sparse convolution~\cite{SubmanifoldSparseConvNet,graham20183d}. PointNet based methods~\cite{qi2017pointnet,qi2018pointnnetplus,thomas2019kpconv} use point-wise MLPs with the ball-query and max-pooling operators to extract features. Sparse convolution based methods~\cite{SubmanifoldSparseConvNet,graham20183d,choy20194d} first convert point clouds to regular grids and handle the sparse volumes efficiently. Though impressive, they suffer from either the intensive computation of sampling and grouping~\cite{qi2018pointnnetplus} or the limited representation capacity due to submanifold dilation~\cite{SubmanifoldSparseConvNet}. More importantly, these specifically-designed operations generally can not be implemented with well-optimized deep learning tools (\emph{e.g.}, PyTorch and TensorFlow) and require writing customized CUDA codes, which greatly limits their deployment in real-world applications. 

Witnessing the success of transformer~\cite{attention} in the 2D domain, numerous attention-based 3D vision methods have been investigated in point cloud processing. However, because of the sparse characteristic of 3D points, the number of non-empty voxels in each local region can vary significantly, which makes directly applying a standard Transformer non-trivial. To efficiently process the attention on sparse data, many approaches rebalance the token number by random sampling~\cite{pan20213d,zhao2021point} or group local regions with similar number of tokens together~\cite{fan2021embracing,sun2022swformer}. These methods are either inseparable from superfluous computations (\textit{e.g.}, dummy token padding and non-parallel attention) or noncompetitive performance (\textit{e.g.}, token random dropping). Alternatively, some approaches~\cite{mao2021voxel,voxelset} try to solve these problems by writing customized CUDA operations, which require heavy optimization to be deployed on modern devices. Hence, building an efficient and deployment-friendly 3D transformer backbone is the main challenge we aim to address in this paper.

In this paper, we seek to expand the applicability of Transformer such that it can serve as a powerful backbone for outdoor 3D perception, as it does in 2D vision. Our backbone is efficient yet deployment-friendly without any customized CUDA operations. To achieve this goal, we present two major modules, one is the dynamic sparse window attention to support efficient parallel computation of local windows with diverse sparsity, and the other one is a novel learnable 3D pooling operation to downsample the feature map and better encode geometric information.  

\begin{figure}[t]
  \centering
   \includegraphics[width=0.9\linewidth]{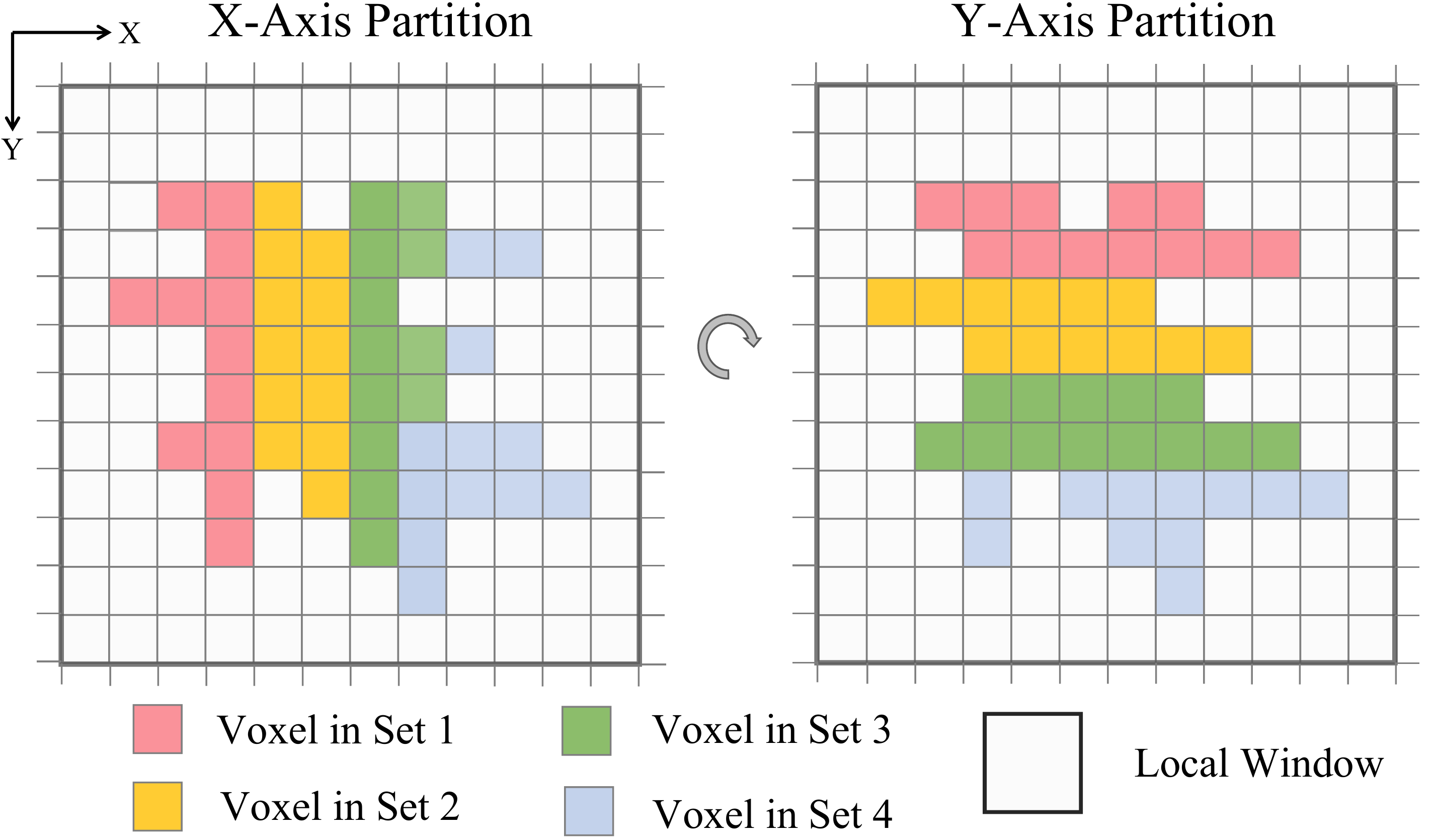}
   \vspace{-2pt}
   \caption{A demonstration of dynamic sparse window attention in our DSVT block. In the X-Axis DSVT layer, the sparse voxels will be split into a series of window-bounded and size-equivalent subsets in X-Axis main order, and self-attention is computed within each set. In the next layer, the set partition is switched to Y-Axis, providing connections among the previous sets.}
   \label{fig:dynamicsetatt}
   \vspace{-16pt}
\end{figure}

Specifically, as illustrated in Figure \ref{fig:dynamicsetatt}, given the sparse voxelized representations and window partition, we first divide each window's sparse voxels into some non-overlapped subsets, where each subset is guaranteed to have the same number of voxels for parallel computation. The partitioning configuration of these subsets will be changed in consecutive self-attention layers based on the rotating partition axis between the X-Axis and Y-Axis. It bridges the subsets of preceding layers for intra-window fusion, providing connections among them that significantly enhance modeling power (see Table \ref{tab:ab_set_part}). To better process the inter-window fusion and encode multi-scale information, we propose the hybrid window partition, which alternates its window shape in successive transformer blocks. It leads to a drop in computation cost with even better performance (see Table \ref{tab:hybridwin}). With the above designs, we process all regions in a fully parallel manner by calculating them in the same batch. This strategy is efficient in regards to real-world latency: i) all the windows are processed in parallel, which is nearly independent of the voxel distribution, ii) using self-attention without key duplication, which facilitates memory access in hardware. Our experiments show that the dynamic sparse window attention approach has much lower latency than previous bucketing-based strategies~\cite{sun2022swformer,fan2021embracing} or vanilla window attention~\cite{liu2021swin}, yet is similar in modeling power (see Table \ref{tab:analyis_att}).

Secondly, we present a powerful yet deployed-friendly 3D sparse pooling operation to efficiently process downsampling and encode better geometric representation. To tackle the sparse characteristic of 3D point clouds, previous methods adopt some custom operations, \textit{e.g.}, customized scatter function~\cite{fan2022fully} or strided sparse convolution to generate downsampled feature volumes~\cite{zhou2018voxelnet,yan2018second}. The requirement of heavy optimization for efficient deployment limits their real-world applications. More importantly, we empirically find that inserting some linear or max-pooling layers between our transformer blocks also harms the network convergence and the encoding of geometric information. 
To address the above limitations, we first convert the sparse downsampling region into dense and process an attention-style 3D pooling operation to automatically aggregate the local spatial features. Our 3D pooling module is powerful and deployed-friendly without any self-designed CUDA operations, and the performance gains (see Table \ref{tab:ab_pooling}) demonstrate its effectiveness.

In a nutshell, our contributions are four-fold: 1) We propose Dynamic Sparse Window Attention, a novel window-based attention strategy for efficiently handling sparse 3D voxels in parallel. 2) We present a learnable 3D pooling operation, which can effectively downsample sparse voxels and encode geometric information better. 3) Based on the above key designs, we introduce an efficient yet deployment-friendly transformer 3D backbone without any customized CUDA operations. It can be easily accelerated by NVIDIA TensorRT to achieve real-time inference speed (\textit{27Hz}), as shown in Figure \ref{fig:speeddemo}. 4) Our approach outperforms previous state-of-the-art methods on the large-scale Waymo~\cite{sun2020scalability} and nuScenes~\cite{caesar2020nuscenes} datasets across various 3D perception tasks.
\section{Related Work} \label{relatedwork} 
\noindent \textbf{3D Perception on Point Clouds.} Previous 3D perception methods can be coarsely classified into point-based methods and voxel-based methods in terms of point representations. Thanks to the powerful PointNet series~\cite{qi2017pointnet,qi2018pointnnetplus}, point-based methods~\cite{shi2019pointrcnn,qi2019deep,wang2022rbgnet,yang20203dssd,cheng2021back,liu2021group} have become extensively employed to extract geometric features directly from raw point clouds. However, these methods suffer from the time-consuming process of point sampling and neighbor searching, and computation-intensive point-wise feature duplication. Voxel-based methods~\cite{yan2018second,shi2020pv,yin2021cvpr,Deng2021VoxelRT,shi2021pv,shi2020p2,wang2022cagroupd,guan2022m3detr,dong2022mssvt} are mainly applied in outdoor autonomous driving scenarios. The input point clouds are first converted into regular 3D voxels and then processed by 3D sparse convolution. Though impressive, to avoid computation expansion, these methods usually adopt submanifold sparse convolution, which greatly decreases the receptive field and limits its representation capacity. More importantly, both the PointNet++ and sparse convolution are implemented with customized CUDA codes, which require heavy optimization for being deployed in real-world applications. To tackle this problem, we propose an efficient yet deployment-friendly point cloud processor for boosting the development of 3D perception.

\noindent \textbf{Transformer for 3D Perception.} Motivated by the significant success of Transformer~\cite{attention} in computer vision community, many researchers have tried to introduce this architecture into point cloud processing. Due to the sparse nature of point clouds, how to efficiently apply a standard transformer is non-trivial. VoTr~\cite{mao2021voxel} first proposes local attention and dilated attention with a self-designed voxel query to enable attention mechanism on sparse voxels. SST~\cite{fan2021embracing} and SWFormer~\cite{sun2022swformer} batch regions with similar number of tokens together and pads them separately to implement parallel computation. Although these methods have explored various sparse attention strategies, they are inevitable to redundant computations (\textit{e.g.} non-parallel attention, token duplication) or uncompetitive performance (\textit{e.g.} random token dropping and sampling). Besides that, most of these methods are also highly dependent on some self-designed CUDA operations (\textit{e.g.}, scatter function~\cite{voxelset} and query function~\cite{mao2021voxel}), which greatly limits their real-world application. To address these problems, we present dynamic sparse window attention, an efficient attention mechanism that computes all the sparse tokens in the same batch in parallel only based on the well-optimized deep learning framework (\textit{e.g.}, PyTorch).

\section{Methodology}
\subsection{Overview}
\begin{figure*}
  \centering
  \includegraphics[width=0.95\linewidth]{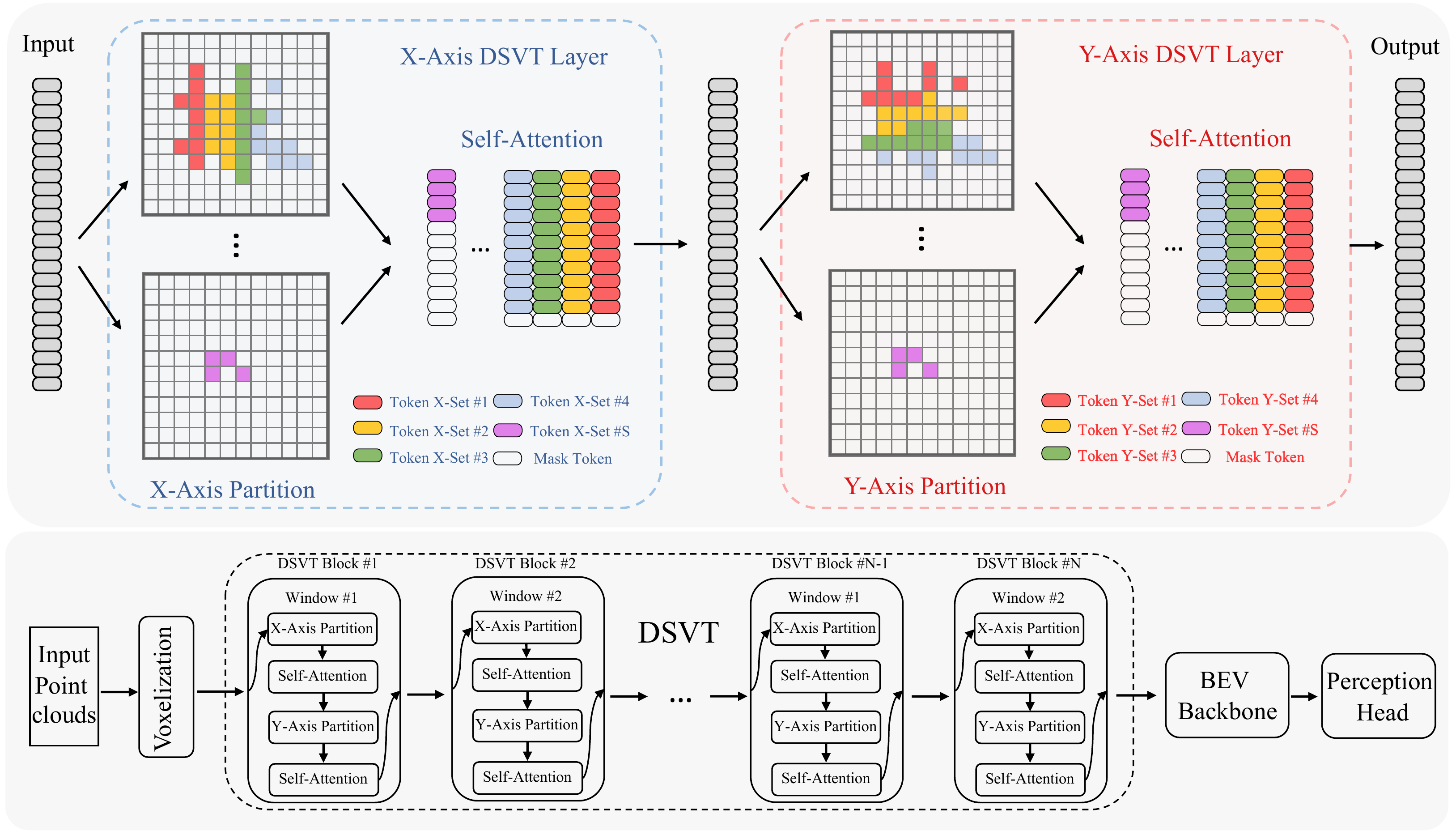}
  \vspace{-6pt}
    \caption{\textbf{Top}: An illustration of the Dynamic Sparse Voxel Transformer block, including one X-Axis DSVT Layer and one Y-Axis DSVT Layer with different set partitions. \textbf{Bottom}: The overall architecture of our proposed DSVT for 3D perception from the point cloud.}
  \label{fig:overview}
  \vspace{-14pt}
\end{figure*}

An overview of the DSVT architecture is presented in Figure \ref{fig:overview}, which illustrates the pillar version (DSVT-P). It first converts the input point clouds into sparse voxels by a voxel feature encoding (VFE) module, like previous voxel-based methods~\cite{wang2020pillar,yin2021cvpr,sun2022swformer}. Each voxel will be treated as a ``token''. Considering the sufficient receptive field of transformer and the tiny scale of outdoor objects, instead of using hierarchical representation, we follow \cite{fan2021embracing} to adopt a single-stride network design, which doesn't reduce the scale of the feature map in X/Y-Axis and is demonstrated to perform better in outdoor 3D Object Detection. With this design, several voxel transformer blocks with dynamic sparse window attention (\textit{DSVT Block}) are applied on these voxels. To provide connections among the sparse voxels, we design two partitioning approaches: rotated set and hybrid window (described in \S \ref{sec:dynaatt}), which introduce intra- and inter-window feature propagation while maintaining efficient computation. To support a 3D variant and better encode accurate 3D geometric information without any extra CUDA operations, a learnable 3D pooling module is designed for effective downsampling (See \S\ref{sec:3Dpool}). Voxel features extracted by our proposed DSVT are then projected to a bird’s eye view (BEV) feature map. In the end, various perception heads can be attached for diverse 3D perception tasks~(See \S\ref{sec:detection}). Our proposed architecture can seamlessly substitute the 3D backbone in existing methods, enhancing outdoor 3D perception performance.

\subsection{Dynamic Sparse Window Attention} \label{sec:dynaatt}
Window-based Transformer architectures
~\cite{liu2021swin,dong2022cswin} have made great success in 2D object detection. However, since point clouds are sparsely distributed in 3D space, many voxels are empty with no valid points. Therefore, the number of non-empty voxels in each window may vary significantly, which makes directly applying a standard Transformer non-trivial. The computation cost of simply padding the sparse volume to dense in each window is expensive (See Table \ref{tab:analyis_att}). Previous methods~\cite{fan2021embracing,voxelset,sun2022swformer} try to rebalance the voxel number by sampling~\cite{zhao2021point,pan20213d} or grouping regions with a similar number of tokens together~\cite{sun2022swformer,fan2021embracing}. Though impressive, they suffer from redundant computations or unstable performance. Alternatively, some approaches write customized CUDA operations to alleviate these problems, which generally have inferior running speed than well-optimized native operations in deep learning frameworks (\emph{e.g.}, PyTorch), limiting their deployment in real-world applications.

To address the above limitations, we propose Dynamic Sparse Window Attention, a window-based attention strategy for efficiently handling sparse 3D voxels in parallel. Notably, it is all implemented by well-optimized native operations in deep learning tools without any self-designed CUDA operations, thus is friendly to be deployed on modern GPUs.

\noindent \textbf{Dynamic set partition.} To efficient perform standard attention among the given sparse voxels inside each window, we reformulate it as parallel computing self-attention within a series of window-bounded and size-equivalent subsets.

Specifically, after converting points to 3D voxels, they are further partitioned into a list of non-overlapping 3D windows with fixed size $L \times W \times H$, like previous window-based approaches~\cite{liu2021swin,sun2022swformer,fan2021embracing}. As for a specific window, it has $N$ non-empty voxels, \textit{i.e.},
\vspace{-4pt}
\begin{equation}
  \mathcal{V} = \{ v_i | v_i= [(x_i, y_i, z_i); f_i; d_i] \}^N_{i=1}, 
  \vspace{-2pt}
\end{equation}
where $(x_i, y_i, z_i) \in \mathbb{R}^3$ and $f_i \in \mathbb{R}^C$ denote the coordinates and features of sparse voxels, respectively. $d_i \in [1, N]$ is the corresponding inner-window voxel ID, which can be generated by specifying a sorting strategy of these voxels. To generate non-overlapped and size-equivalent local sets, 
we first compute the required number of sub-sets in this window, as follows: 
\vspace{-4pt}
\begin{equation}
    S = \lfloor \frac{N}{\tau} \rfloor + \mathbb{I}[(N~\%~\tau) > 0],
    \vspace{-2pt}
\end{equation}
where $\lfloor \cdot \rfloor$ is the floor function and $\mathbb{I}[\cdot]$ is the indicator function. $\tau$ is a hyperparameter that indicates the maximum number of non-empty voxels allocated to each set. In this way, we can cover all the voxels in this window with a minimum number of subsets. Notably, $S$ dynamically varies with the sparsity of the window. The more non-empty voxels, the more sets and computation resources will be assigned to process this window, which is the key design of our dynamic sparse window attention. 

With the number of assigned sets $S$, we  
evenly distribute $N$ non-empty voxels into $S$ sets.  
Specifically, for the voxel indices that belong to $j$-th set (denoted as $\mathcal{Q}_j = \{q^j_k\}^{\tau-1}_{k=0}$),  we compute its $k$-th index as
\vspace{-2pt}
\begin{equation}
  q^j_k = \left\lfloor\frac{ (j \times \tau + k)}{S \times \tau} \times N \right\rfloor,  ~~\text{for}~~ k = 0, ..., \tau - 1,  
  \label{eq:index_calculate}
  \vspace{-2pt}
\end{equation}
where $\lfloor \cdot \rfloor$ is the floor function. 
Note that this operation can generate a specific number (\emph{i.e.}, $\tau$) of voxels for each set, regardless of $N$, which enables the attention layer can be performed on all sets in a fully parallel manner.
Although Eq.~\eqref{eq:index_calculate} may duplicate some voxels within each set to achieve the number of $\tau$ voxels, these redundant voxels will be masked when conducting attention. 

After obtaining the partition $\mathcal{Q}_j$ of $j$-th set, we then collect the corresponding voxel features and coordinates based on the voxel inner-window id $\mathcal{D} = \{d_i\}_{i=1}^N$, as follows,
\vspace{-2pt}
\begin{equation}
  \mathcal{F}_j, \mathcal{O}_j = \text{INDEX}(\mathcal{V}, \mathcal{Q}_j, \mathcal{D}),  
  \label{eq:index}
  \vspace{-2pt}
\end{equation}
where $\text{INDEX}(\cdot _{\text{voxels}}, \cdot _{\text{partition}}, \cdot _{\text{ID}})$ is the index operation, $\mathcal{F}_j \in \mathbb{R}^{\tau \times C}$ and $\mathcal{O}_j \in \mathbb{R}^{\tau \times 3} $ are the corresponding voxel features and spatial coordinates  $(x, y, z)$ of this set. 

In this way, we obtain some non-overlapped and window-bounded subsets with the same number of sparse voxels. Notably, our dynamic set partition is highly dependent on the inner-window voxel ID (see Eq.~\eqref{eq:index}), so we can easily control  
the covered local region of each set by voxel ID reordering with different sorting strategies. 

\noindent \textbf{Rotated set attention for intra-window feature propagation.} 
The above dynamic set partition can group sparse voxels into non-overlapped and window-bounded subsets, where each subset has the same number of voxels that enables to conduct self-attention layers within each subset in a fully parallel manner. 
However, computing self-attention inside the invariant partition lacks connections across the subsets, limiting its modeling power. To bridge the voxels among the non-overlapping sets while maintaining efficient computation, we propose the rotated-set attention approach that alternates between two partitioning configurations in consecutive attention layers.

As illustrated in Figure \ref{fig:overview}, our DSVT block contains two self-attention layers. And the first layer uses the X-Axis Partition, in which the voxel ID is sorted according to their coordinates in X-Axis main order. The next layer adopts a rotated partition configuration that is sorted in Y-Axis main order. With the above design, DSVT block is computed as
\vspace{-4pt}
\begin{equation}
    \begin{aligned}
    &\mathcal{F}^{l}, \mathcal{O}^{l} = \text{INDEX}(\mathcal{V}^{l-1}, \{\mathcal{Q}_j\}_{j=0}^{S-1}, \mathcal{D}_x), \\
    &\mathcal{V}^{l} = \text{MHSA}(\mathcal{F}^{l}, \text{PE}(\mathcal{O}^{l})), \\
    &\mathcal{F}^{l+1}, \mathcal{O}^{l+1} = \text{INDEX}(\mathcal{V}^{l}, \{\mathcal{Q}_j\}_{j=0}^{S-1}, \mathcal{D}_y), \\
    &\mathcal{V}^{l+1} = \text{MHSA}(\mathcal{F}^{l+1}, \text{PE}(\mathcal{O}^{l+1})), 
    \end{aligned}
    \vspace{-2pt}
\end{equation}
where $\mathcal{D}_x$ and $\mathcal{D}_y$ are the inner-window voxel index sorted in X-Axis and Y-Axis respectively. $\text{MHSA}(\cdot)$ denotes the conventional Multi-head Self-Attention layer with FFN and Layer Normalization. $\text{PE}(\cdot)$ stands for the positional encoding function used in \cite{liu2021swin}. $\mathcal{F} \in \mathbb{R}^{S \times \tau \times C}$ and $\mathcal{O} \in \mathbb{R}^{S \times \tau \times 3}$ are the corresponding indexed voxel features and coordinates of all sets. To this end, the original sparse window attention will be approximatively reformulated as several set attention, which can be processed in the same batch in parallel. In this way, it is easy to implement an efficient DSVT module in current popular deep learning frameworks without extra engineering efforts of deployment.

The rotated set partition introduces connections between non-overlapping sets in the previous layer and is found to be effective in outdoor 3D perception~(see Table \ref{tab:ab_set_part}).

\noindent \textbf{Hybrid window partition for inter-window feature propagation.} To connect the voxels across windows and reduce the computation cost, we present the hybrid window partition approaches at the block level. We observe that our dynamic window attention layer has linear computational complexity with respect to the total number of subsets. However, increasing the window sizes will reduce the number of sets and lead to the drop of self-attention's computation cost, but also destroy the detection performance for small objects. This motivates us to adopt the hybrid window splitting to deliver good performance-efficiency trade-off. 

Specifically, we follow the swin-transformer~\cite{liu2021swin} using a window shifting technique between two consecutive DSVT blocks to re-partition the sparse window, but their window size is different. With this design, our block can effectively save computation costs without sacrificing performance.

\subsection{Attention-style 3D Pooling} \label{sec:3Dpool}\vspace{-3pt}
Compared with the pillar-style 2D encoder, the voxel-based 3D backbone can capture more precise position information, which is beneficial for 3D perception. However, we observed that simply padding the sparse regions and applying an MLP network for downsampling as \cite{liu2021swin} will drop the performance (See Table \ref{tab:ab_pooling}). It is challenging for one-layer MLP to learn the sparse geometric information with too many zero-padded holes. Moreover, we observe that inserting an MLP between our transformer blocks also harms network optimization. To support effective 3D downsampling and better encode spatial information, we present an attention-style 3D pooling operation, which is compatible with our full attention setting and implemented without any self-design CUDA operations.

Given a pooling local region $l \times w \times h$, (\textit{e.g.}, $2 \times 2 \times 2$), with $P$ non-empty voxels $\{p_i\}_{i=1}^{P}$, we first pad this sparse region into dense, $\{\widetilde{p}_i\}_{i=1}^{l \times w \times h}$, and perform standard max-pooling along the voxel dimension as 
\vspace{-2pt}
\begin{equation}
    \mathcal{P} = \text{MaxPool}(\{\widetilde{p}_i\}_{i=1}^{P}).
    \vspace{-2pt}
\end{equation}
Instead of directly feeding the pooled feature to the succeeding module, we use pooled feature $\mathcal{P}$ to construct the query vector, while the original unpooled $\{\widetilde{p}_i\}_{i=1}^{P}$ serves that role of key and value vectors, \textit{i.e.},
\vspace{-2pt}
\begin{equation}
   \widetilde{\mathcal{P}} = \text{Attn}(\text{Q}=\mathcal{P}, \text{KV}=\{\widetilde{p}_i\}_{i=1}^{l \times w \times h})
   \vspace{-2pt}
\end{equation}
With this attention-style 3D pooling operation, our 3D backbone holds the characteristic of fully attention and achieves better performance than our pillar variant.

\begin{table*}
  \vspace{-5pt}
  \centering
  \resizebox{\textwidth}{!}{
  \begin{tabular}{l | c | c |>{\columncolor{mygray}} c| >{\columncolor{mygray}} c|c|c|c|c|c|c}
    \toprule
    \toprule
    & & & mAP/mAPH & mAP/mAPH & \multicolumn{2}{c|}{\textit{Vehicle} 3D AP/APH} & \multicolumn{2}{c|}{\textit{Pedestrian} 3D AP/APH} & \multicolumn{2}{c}{\textit{Cyclist} 3D AP/APH} \\
    \multirow{-2}{*}{Methods} & \multirow{-2}{*}{Present at} &\multirow{-2}{*}{Stages}   & L1 & L2   & L1 & L2 & L1 & L2 & L1 & L2 \\
    \midrule
    SECOND~\cite{yan2018second} & Sensors'18 & One &67.2/63.1 & 61.0/57.2 &72.3/71.7 & 63.9/63.3 & 68.7/58.2 & 60.7/51.3 & 60.6/59.3 & 58.3/57.0 \\
    PointPillars\ddag~\cite{lang2019pointpillars} & CVPR'19 &One &69.0/63.5 & 62.8/57.8 & 72.1/71.5 & 63.6/63.1 & 70.6/56.7 & 62.8/50.3 & 64.4/62.3 & 61.9/59.9 \\
    CenterPoint-Voxel\dag~\cite{yin2021cvpr} &CVPR'21&One & 74.4/71.7 & 68.2/65.8 & 74.2/73.6& 66.2/65.7 & 76.6/70.5 & 68.8/63.2 & 72.3/71.1 & 69.7/68.5 \\
    SST\ddag~\cite{fan2021embracing} &CVPR'22&One &74.5/71.0 & 67.8/64.6 & 74.2/73.8 & 65.5/65.1 & 78.7/69.6 & 70.0/61.7 & 70.7/69.6  & 68.0/66.9 \\
    VoxSet~~\cite{voxelset} &CVPR'22& One &75.4/72.2 & 69.1/66.2 & 74.5/74.0 & 66.0/65.6 & 80.0/72.4 & 72.5/65.4 & 71.6/70.3 & 69.0/67.7 \\
    AFDetV2~\cite{hu2022afdetv2} & AAAI'22&One &77.2/74.8 & 71.0/68.8 & 77.6/77.1 & 69.7/69.2 & 80.2/74.6 & 72.2/67.0 & 73.7/72.7  & 71.0/70.1 \\
    SWFormer~\cite{sun2022swformer} &ECCV'22& One &-/-  & -/-       &77.8/77.3 & 69.2/68.8 & 80.9/72.7 & 72.5/64.9 & -/- & -/- \\
    PillarNet-34~\cite{shi2022pillarnet}&ECCV'22 &One &77.3/74.6 & 71.0/68.5 & 79.1/78.6 & 70.9/70.5 & 80.6/74.0 & 72.3/66.2 & 72.3/71.2  & 69.7/68.7 \\
    CenterFormer~\cite{Zhou_centerformer} &ECCV'22&One &75.3/72.9 &71.1/68.9 & 75.0/74.4 & 69.9/69.4 & 78.6/73.0 & 73.6/68.3& 72.3/71.3 & 69.8/68.8 \\
    Ours (Pillar) & - &One &\textbf{79.5/77.1} & \textbf{73.2/71.0} & \textbf{79.3/78.8} & \textbf{70.9/70.5} & \textbf{82.8/77.0} & \textbf{75.2/69.8} & \textbf{76.4/75.4} & \textbf{73.6/72.7} \\
    Ours (Voxel)& - &One &\textbf{80.3/78.2} & \textbf{74.0/72.1} & \textbf{79.7/79.3} & \textbf{71.4/71.0} & \textbf{83.7/78.9} & \textbf{76.1/71.5} & \textbf{77.5/76.5} & \textbf{74.6/73.7} \\
    \midrule
    PV-RCNN\dag~\cite{shi2020pv} &CVPR'20&Two& 76.2/73.6 & 69.6/67.2 & 78.0/77.5 & 69.4/69.0 & 79.2/73.0 & 70.4/64.7 & 71.5/70.3 & 69.0/67.8 \\
    Part-A2-Net~\cite{shi2020p2} &TPAMI'20 &Two& 73.6/70.3 & 66.9/63.8 & 77.1/76.5 & 68.5/68.0 & 75.2/66.9 & 66.2/58.6 & 68.6/67.4 & 66.1/64.9 \\
    CenterPoint-Voxel~\cite{yin2021cvpr} &CVPR'21&Two& -/- & -/- & 76.7/76.2 & 68.8/68.3 & 79.0/72.9 & 71.0/65.3 & -/- & -/- \\
    PV-RCNN++(center)~\cite{shi2021pv} &IJCV'22&Two& 78.1/75.9 & 71.7/69.5 & 79.3/78.8 & 70.6/70.2 & 81.3/76.3 & 73.2/68.0 & 73.7/72.7  & 71.2/70.2 \\
    FSD~\cite{fan2022fully} &NeurIPS'22&Two& 79.6/77.4 & 72.9/70.8 & 79.2/78.8 & 70.5/70.1 &82.6/77.3 &73.9/69.1 &77.1/76.0 &74.4/73.3 \\
    Ours (Pillar-TS) & - &Two&\textbf{80.6/78.2} & \textbf{74.3/72.1} & \textbf{80.2/79.7} & \textbf{72.0/71.6} & \textbf{83.7/78.0} & \textbf{76.1/70.7} & \textbf{77.8/76.8} & \textbf{74.9/73.9} \\
    Ours (Voxel-TS)  & - &Two& \textbf{81.1/78.9} & \textbf{74.8/72.8} & \textbf{80.4/79.9} & \textbf{72.2/71.8} & \textbf{84.2/79.3} & \textbf{76.5/71.8} & \textbf{78.6/77.6} & \textbf{75.7/74.7} \\
    \midrule
    \midrule
    CenterFormer(2f)~\cite{Zhou_centerformer} &ECCV'22 &One &78.3/76.7 &74.3/72.8 & 77.0/76.5 & 72.1/71.6  & 81.4/78.0 & 76.7/73.4& 76.6/75.7 &  74.2/73.3 \\
    Ours (Pillar-2f) & - &One&\textbf{81.4/79.8} & \textbf{75.4/73.9} & \textbf{80.8/80.3} & \textbf{72.7/72.3} & \textbf{84.5/81.3} & \textbf{77.2/74.1} & \textbf{78.8/77.9} & \textbf{76.3/75.4} \\
    Ours (Voxel-2f)  & - &One& \textbf{81.9/80.4} & \textbf{76.0/74.6} & \textbf{81.1/80.6} & \textbf{73.0/72.6} & \textbf{84.9/81.7} & \textbf{77.8/74.8} & \textbf{79.8/78.9} & \textbf{77.3/76.4} \\
    \midrule
    SST(3f)~\cite{fan2021embracing} &CVPR'22 &One &-/- & -/- & 77.0/76.6  & 68.5/68.1 & 82.4/78.0 & 75.1/70.9 & -/-  & -/- \\
    Ours (Pillar-3f) & - &One&\textbf{81.9/80.5} & \textbf{76.2/74.8} & \textbf{81.2/80.8} & \textbf{73.3/72.9} & \textbf{85.0/82.0} & \textbf{78.0/75.0} & \textbf{79.6/78.8} & \textbf{77.2/76.4} \\
    Ours (Voxel-3f) & - &One&\textbf{82.1/80.8} & \textbf{76.3/75.0} & \textbf{81.5/81.1} & \textbf{73.6/73.2} & \textbf{85.3/82.4} & \textbf{78.2/75.4} & \textbf{79.6/78.8} & \textbf{77.2/76.4} \\
    \midrule
    CenterFormer(4f)~\cite{Zhou_centerformer}&ECCV'22 &One &78.5/77.0 &74.7/73.2 & 78.1/77.6 &  73.4/72.9  & 81.7/78.6 &  77.2/74.2& 75.6/74.8 &  73.4/72.6 \\
    Ours (Pillar-4f) & - &One&\textbf{82.5/81.0} & \textbf{76.7/75.3} & \textbf{81.7/81.2} & \textbf{73.8/73.4} & \textbf{85.4/82.3} & \textbf{78.5/75.5} & \textbf{80.3/79.4} & \textbf{77.9/77.1} \\
    Ours (Voxel-4f) & - &One&\textbf{82.6/81.3} & \textbf{76.9/75.6} & \textbf{81.8/81.4} & \textbf{74.1/73.6} & \textbf{85.6/82.8} & \textbf{78.6/75.9} & \textbf{80.4/79.6} & \textbf{78.1/77.3} \\
    \midrule
    \midrule
    PV-RCNN++(2f)\dag~\cite{shi2021pv}  &IJCV'22&Two &79.4/78.0 & 73.3/71.9& 80.2/79.7 & 72.1/71.7 & 83.5/80.4 & 75.5/72.6 & 74.6/73.8 & 72.4/71.5 \\
    Ours (Pillar-TS-2f) & - &Two&\textbf{81.9/80.4} & \textbf{76.0/74.5} & \textbf{81.3/80.9} & \textbf{73.4/73.0} & \textbf{85.2/81.9} & \textbf{77.9/74.7} & \textbf{79.2/78.3} & \textbf{76.7/75.9} \\
    Ours (Voxel-TS-2f) & - &Two&\textbf{82.3/80.8} & \textbf{76.6/75.1} & \textbf{81.4/81.0} & \textbf{73.5/73.1} & \textbf{85.4/82.2} & \textbf{78.4/75.3} & \textbf{80.2/79.3} & \textbf{77.8/76.9} \\
    \midrule
    SST\_TS(3f)~\cite{fan2021embracing}  &CVPR'22&Two &-/- & -/-& 78.7/78.2 & 70.0/69.6 & 83.8/80.1 & 75.9/72.4 & -/- & -/- \\
    Ours (Pillar-TS-3f) & - &Two&\textbf{82.5/81.0} & \textbf{76.7/75.4} & \textbf{81.8/81.3} & \textbf{74.0/73.6} & \textbf{85.6/82.6} & \textbf{78.5/75.6} & \textbf{80.1/79.2} & \textbf{77.7/76.9} \\
    Ours (Voxel-TS-3f) & - &Two&\textbf{82.6/81.2} & \textbf{76.8/75.5} & \textbf{81.8/81.4} & \textbf{74.0/73.6} & \textbf{85.8/82.9} & \textbf{78.8/75.9} & \textbf{80.1/79.2} & \textbf{77.7/76.9} \\
    \midrule
    MPPNet~(4f) \cite{chen2022mppnet} &ECCV'22 &Two& 81.1/79.9 &75.4/74.2 & 81.5/81.1 &   74.1/73.6 &84.6/82.0  & 77.2/74.7 & 77.2/76.5 &  75.0/74.4 \\
    Ours (Pillar-TS-4f) & - &Two&\textbf{82.9/81.5} & \textbf{77.3/75.9} & \textbf{82.1/81.6} & \textbf{74.4/74.0} & \textbf{85.8/82.8} & \textbf{79.0/76.1} & \textbf{80.9/80.0} & \textbf{78.6/77.7} \\
    Ours (Voxel-TS-4f) & - &Two&\textbf{83.1/81.7} & \textbf{77.5/76.2} & \textbf{82.1/81.6} & \textbf{74.5/74.1} & \textbf{86.0/83.2} & \textbf{79.1/76.4} & \textbf{81.1/80.3} & \textbf{78.8/78.0} \\
    \toprule
  \end{tabular}
  }
  \vspace{-8pt}
  \caption{The results of 3D object detection on Waymo Open validation set (100\% training data). ``2f'', ``3f'' and ``4f'' stand for 2-, 3- and 4-frame models.~\dag: re-implemented by OpenPCDet. \ddag: reported by ~\cite{fan2022fully}. We highlight the top-2 entries with \textbf{bold} font in each column. }
  \label{tab:one_results}
  \vspace{-12pt}
\end{table*}

\begin{table}
  \vspace{-5pt}
  \centering
  \resizebox{\linewidth}{!}{
  \begin{tabular}{l|c|c|c|c|c}
    \toprule
    \toprule
    &  & \multicolumn{2}{c|}{\textit{val}} & \multicolumn{2}{c}{\textit{test} } \\
    \multirow{-2}{*}{Methods} & \multirow{-2}{*}{Present at} & NDS & mAP & NDS & mAP \\
    \midrule
    PointPillars~\cite{lang2019pointpillars} & CVPR'19 & - & - &45.3 & 30.5 \\
    CBGS~\cite{zhu2019class} & ArXiv'19 &62.3 & 50.6 & 63.3 & 52.8  \\
    CenterPoint-Voxel~\cite{yin2021cvpr} &CVPR'21 & 66.8 & 59.6 & 67.3 & 60.3  \\
    Transfusion-L~\cite{bai2022transfusion} &CVPR'22 & 69.3 & 64.7 &70.2 & 65.5  \\
    PillarNet-34~\cite{shi2022pillarnet}& ECCV'22 & - & - &71.4 & 66.0 \\
    Ours (Pillar) & - & \textbf{71.1} & \textbf{66.4} & \textbf{72.7} & \textbf{68.4} \\
    \toprule
  \end{tabular}
  }
  \vspace{-8pt}
  \caption{The results of 3D object detection on nuScenes Dataset.}
  \label{tab:nusdet_results}
  \vspace{-12pt}
\end{table}

\subsection{Perception Model} \label{sec:detection}\vspace{-3pt}
\noindent \textbf{Backbone Variants} We build our base backbone model, called DSVT-P, a single-stride pillar-based sparse voxel backbone that has a similar model size to the widely-used sparse convolution backbone~\cite{zhou2018voxelnet,yan2018second}. We also introduce DSVT-V, a single-stride voxel-based 3D backbone that only gradually reduces the size of the feature map in the Z-Axis by our attention-style 3D pooling module. Notably, the output voxel number and corresponding coordinates of these two variants are equal, which can be easily used to ablate the effectiveness of our pooling module.

\noindent \textbf{Detection head and Training Objectives.}
Our DSVT is flexible and can be applied in most voxel-based detection frameworks~\cite{yin2021cvpr,lang2019pointpillars}. In the Waymo variant, we follow the framework of CenterPoint-Pillar~\cite{yin2021cvpr} and append our DSVT before BEV backbone. For nuScenes, we simply replace the 3D encoder of Transfusion-L~\cite{bai2022transfusion} with DSVT. Besides that, we also follow \cite{shi2022pillarnet,liang2019multi,YihanHu2022AFDetV2RT} that uses IoU-rectification scheme to incorporate the IoU information into confidence scores. 

\noindent \textbf{Two-stage model.}
Although our main contributions focus on the design of the sparse voxel transformer architecture in the backbone network, there is still a considerable gain between one-stage and two-stage detectors. To match the performance gap, we provide a two-stage version with CT3D~\cite{sheng2021improving}, which is a widely used two-stage framework that performs proposal-aware aggregation from point clouds.

\noindent \textbf{Segmentation head.} We follow the lidar branch of BEVFusion~\cite{liu2022bevfusion} in map segmentation task and only switch the 3D backbone while keeping other components unchanged.
\section{Experiments}
\noindent \textbf{Waymo Open}~\cite{sun2020scalability} is a widely used outdoor 3D perception benchmark consisting of 1150 point cloud sequences in total (more than 200K frames). Each frame covers a large perception range~($150\text{m} \times 150\text{m}$). All results are evaluated by the standard protocol using 3D mean Average Precision (mAP) and its weighted variant by heading accuracy (mAPH).
\begin{table}
  \vspace{-5pt}
  \centering
  \resizebox{\linewidth}{!}{
  \begin{tabular}{l|c|c|c|c|c|c|>{\columncolor{mygray}}c}
    \toprule
    \toprule 
    Encoder & DA & PC & WW & SL & CP & DI & mIoU \\
    \midrule
    2D Conv~\cite{liu2022bevfusion} & 72.0 & 43.1 & 53.1 & 29.7 & 27.7 & 37.5 & 43.8 \\
    3D SpConv~\cite{liu2022bevfusion}  & 75.6 & 48.4 & 57.5 & 36.5 & 31.7 & 41.9 & 48.6 \\
    Ours~(Pillar) &\textbf{79.7}  & \textbf{51.8} & \textbf{61.1} & \textbf{38.2} & \textbf{33.8} & \textbf{45.3} & \textbf{51.6} \\
    Ours~(Pillar)$^\dag$ & \textbf{87.6} & \textbf{67.2} & \textbf{72.7} & \textbf{59.7} & \textbf{62.7} & \textbf{58.2}  & \textbf{68.0} \\
    \toprule
  \end{tabular}
  }
  \vspace{-8pt}
  \caption{The results of BEV map segmentation on nuScenes(val).  All the baselines are reported by \cite{liu2022bevfusion}. 1$^{st}$-3$^{rd}$ rows adopt the same segmentation framework except for 3D encoder. $\dag$ means a deeper BEV backbone. Notion: Drivable(DA), Ped.Cross.(PC), Walkway(WW), StopLine(SL), Carpark(CP), Divider(DI).}
  \label{tab:nusseg_results}
  \vspace{-14pt}
\end{table}

\noindent \textbf{NuScenes}~\cite{caesar2020nuscenes} is a challenging outdoor dataset providing diverse annotations for various tasks, (\emph{e.g.}, 3D object detection and BEV map segmentation). It contains 40,157 annotated samples, each with six cameras and a 32-beam LiDAR scan. For 3D object detection, we report the nuScenes detection score (NDS) and mean average precision (mAP), while the mean IoU is provided for the map segmentation task.
\vspace{-6pt}
\subsection{Implementation Details.} \label{sec:imple}
\vspace{-4pt}
\noindent \textbf{Network Architecture.} For the DSVT-P variant, we build our backbone with 4 blocks and each block consists of two DSVT attention layers. In case of Waymo 3D object detection, we use the grid size of (0.32m, 0.32m, 6m) following CenterPoint-Pillar~\cite{yin2021cvpr}. The hybrid window sizes are set to (12, 12, 1) and (24, 24, 1), and the maximum number of voxels assigned to each set is 36. For the DSVT-V, this variant is similar to the pillar-based framework except for splitting the Z-Axis into more voxels. Specifically, its backbone has four stages with block numbers \{1, 1, 1, 1\}. The voxel size is held as (0.32m, 0.32m, 0.1875m) and we downsample the voxel feature map using our attention-style 3D pooling module only in the Z-Axis with the stride \{4, 4, 2\}. The window sizes along the Z-Axis are \{32, 8, 2, 1\}, and the maximum number of subsets is set to 48. All the attention modules are 192 input channels. For the nuScenes variant, the architecture differs slightly. More details of architecture and hyper-parameter analysis are in Appendix.\\
\noindent \textbf{Training and Inference.} All the variants are trained by AdamW optimizer~\cite{loshchilov2018decoupled} on 8 NVIDIA A100 GPUs. We adopt the same learning rate scheme as \cite{yin2021cvpr}. All inference times are profiled on the same workstation (single NVIDIA A100 GPU). More implementation details are in Appendix.

\begin{table}
\begin{center}
    \centering
        \resizebox{0.95\linewidth}{!}{
        \begin{tabular}{c|cccc}
        \Xhline{0.8pt}\noalign{\smallskip}
        Backbone &\#param. & Latency & L2 mAP& L2 mAPH\\
        \noalign{\smallskip}
        \hline
        \noalign{\smallskip}
        ResBackbone1x & 88M &56ms & 69.61 &66.81\\
        DSVT(Pillar) & 71M &60ms &\textbf{71.14} &\textbf{68.59}\\
        \Xhline{0.8pt}
        \end{tabular}
        }
\end{center}
\vspace{-16pt}
\caption{Comparison with sparse convolution. Only switch the 3D backbone while other components remain unchanged.}
\label{table:backbone_compare}
\vspace{-10pt}
\end{table}

\begin{table}
\begin{center}
    \centering
        \resizebox{\linewidth}{!}{
        \begin{tabular}{lccccc}
        \Xhline{0.8pt}\noalign{\smallskip}
        Method & Latency & PadRatio &Memory  & L\_2 mAP& L\_2 mAPH\\
        \noalign{\smallskip}
        \hline
        \noalign{\smallskip}
        Fully-Padding & 77ms & - &7974MB & 64.17 & 60.94 \\
        Bucketing~\cite{fan2021embracing}& 52ms & 35.1\% &3566MB & 64.07& 60.86 \\
        Ours (12 $\times$ 12) & 30ms  & 38.9\% & 3688MB &\textbf{64.41}&\textbf{61.22} \\
        Ours (12 $\times$ 24) & 29ms  & 28.3\% & 3640MB &\textbf{64.59}&\textbf{61.40} \\
        \Xhline{0.8pt}
        \end{tabular}
        }
\end{center}
\vspace{-16pt}
\caption{Comparison with standard self-attention. We report the latency of transformer backbone based on the code base of SST~\cite{fan2021embracing}.}
\label{tab:analyis_att}
\vspace{-16pt}
\end{table}

\subsection{3D Object Detection} \label{sec:3DObjDet}
For performance benchmarking, we compare our model with previous methods on Waymo and NuScenes datasets. \\
\noindent \textbf{Waymo.} All the results are summarized in Table \ref{tab:one_results}. As for the one-stage framework, our single frame DSVT-V model achieves 72.1 mAPH on L2 difficulties, which is +3.2 better than the previous best one-stage method~\cite{Zhou_centerformer} and even +1.3 higher than the best two-stage approaches~\cite{fan2022fully}. Compared with our pillar variant, DSVT-V obtains significant improvements in small objects, (\textit{i.e.}, pedestrian and cyclist), which demonstrates the effectiveness of our 3D pooling module for capturing detailed geometric information. Moreover, our model reaches 74.6, 75.6 in terms of L2 mAPH on 2, 4 frame settings, which outperforms the previous best one-stage multi-frame methods~\cite{Zhou_centerformer} by +1.8 and +2.4, respectively. These gains are more remarkable on L1 mAPH, \textit{i.e.}, +3.7 and +4.3. Note that our model is not specifically designed for multi-frame detection (\textit{e.g.}, multi-frame fusion), and only takes concatenated point clouds as input. 

We also evaluate our model against several competing two-stage approaches in Table~\ref{tab:one_results}. Our DSVT-TS voxel variant outperforms all the previous methods on all frame settings with a large margin, which gains +2.0, +3.2 and +2.0 of 3D mAPH(LEVEL\_2) on 1-, 2- and 4-frame settings respectively. Note that we simply use raw-point based CT3D~\cite{sheng2021improving} as our second stage without some other elaborate designs, (\textit{i.e.}, features interpolated from feature maps), leaving huge space for exploring better two-stage performance.

\noindent \textbf{NuScenes.} As shown in Table \ref{tab:nusdet_results}, our model outperforms all methods with remarkable gains. It achieves state-of-the-art performance, 72.7 and 68.4 in terms of \textit{test} NDS and mAP, surpassing PillarNet~\cite{shi2022pillarnet} by +1.3 and +2.4 separately.

\subsection{BEV map segmentation} \label{sec:mapseg}
To further demonstrate the generalizability of our approach, we evaluate our model on BEV Map Segmentation task of nuScenes dataset with only lidar input. The comparison results with state-of-the-art methods are reported in Table \ref{tab:nusseg_results}. Note that we follow the lidar branch of BEVFusion~\cite{liu2022bevfusion} and only switch the 3D backbone. All the baselines are reported by \cite{liu2022bevfusion}. Thanks to the large receptive field of Transformer, our DSVT can further boost the performance with a remarkable gain (+3.0). With a deeper bev backbone, our model can be further improved by a huge gain~(+19.4). 

\begin{table}
\begin{center}
    \centering
        \resizebox{0.95\linewidth}{!}{
        \begin{tabular}{cccc|cc}
        \Xhline{0.8pt}\noalign{\smallskip}
        Random &Sparse  & Regional & Rotate & L2 mAP& L2 mAPH\\
        \noalign{\smallskip}
        \hline
        \noalign{\smallskip}
        $\checkmark$&   & & &63.66&60.20\\
        & $\checkmark$ & &  &64.15&60.89\\
        &  & $\checkmark$&  &64.27&61.03\\
        &  & & $\checkmark$ &\textbf{64.59} &\textbf{61.40}\\
        \Xhline{0.8pt}
        \end{tabular}
        }
\end{center}
\vspace{-16pt}
\caption{Effect of set partition method. All the experiments are based on the code base of SST~\cite{fan2021embracing}.}
\label{tab:ab_set_part}
\vspace{-16pt}
\end{table}

\subsection{Ablation Studies} \label{sec:ablation}
In this section, a set of ablative studies are conducted on Waymo \textit{val} to investigate the key designs of our model. We follow OpenPCDet~\cite{openpcdet2020} to train all models on 20\% training data with 30 epochs. Notably, all the variants are dim-128.

\noindent\textbf{Comparison with sparse convolution backbone.} Following the VoxelBackBone8x used in CenterPoint-Voxel~\cite{yin2021cvpr}, we extend it to a single-stride pillar-based residual backbone with a similar model size of our DSVT-P, named ResBackBone1x. For a fair comparison, we only switch the sparse 3D backbone while all other settings remain unchanged. Note that this variant is a very strong baseline, which is +4.9 higher than the original version~\cite{yin2021cvpr} on L2 mAPH. As evidenced in $1^{st}$ and $2^{nd}$ rows of Table~\ref{table:backbone_compare}, thanks to the large receptive field of Transformer, even on such a strong baseline, our DSVT-P still brings +1.78 L2 mAPH gains over sparse convolution with a slightly larger latency. Due to the characteristic of friendly deployment, our model can achieve 2 $\times$ faster by TensorRT acceleration. More details of the architecture and analysis are in Appendix.

\begin{table}
\begin{center}
    \centering
        \resizebox{0.95\linewidth}{!}{
        \begin{tabular}{ccccc}
        \Xhline{0.8pt}\noalign{\smallskip}
        Window1 & Window2 & Latency & L2 mAP& L2 mAPH\\
        \noalign{\smallskip}
        \hline
        \noalign{\smallskip}
        (12, 12)&(12, 12) &6.04ms& 70.93 & 68.35\\
        (12, 12)&(24, 24) &5.79ms& \textbf{71.14} & \textbf{68.59}\\
        (12, 12)&(36, 36) &\textbf{5.77ms}& 70.80 & 68.18\\
        \Xhline{0.8pt}
        \end{tabular}
        }
\end{center}
\vspace{-16pt}
\caption{Effect of hybrid window partition. The reported latencies are the consumption of one DVST block.}
\label{tab:hybridwin}
\vspace{-10pt}
\end{table}
\begin{table}
\begin{center}
    \centering
        \resizebox{0.7\linewidth}{!}{
        \begin{tabular}{ccc}
        \Xhline{0.8pt}\noalign{\smallskip}
        Pooling Module & L\_2 mAP & L\_2 mAPH\\
        \noalign{\smallskip}
        \hline
        \noalign{\smallskip}
        Pillar & 71.14 & 68.59 \\
        Linear& 70.91&68.60\\
        Max Pooling & 71.21& 68.91 \\
        Attention+Mask& 71.24&68.94  \\
        Attention& \textbf{71.65}& \textbf{69.31}\\
        \Xhline{0.8pt}
        \end{tabular}
        }
\end{center}
\vspace{-15pt}
\caption{Effect of 3D sparse pooling module. The set size of pillar- and voxel-based variants are unified to 36 for a fair comparison.}
\label{tab:ab_pooling}
\vspace{-18pt}
\end{table}

\noindent\textbf{Effect of dynamic sparse window attention.} Table~\ref{tab:analyis_att} ablates the modeling power and efficiency of our dynamic sparse window attention. All the experiments are conducted on the code base of SST~\cite{fan2021embracing}, which applies a bucketing-based method~\cite{sun2022swformer,fan2021embracing} to group the windows into several batches for efficient computation. Fully-padding means that simply converts all the non-empty sparse windows into dense ones. Notably, we eliminate the influence of window shift and only switch the attention strategy used in sparse backbone. Compared with the previous methods, the transformer backbone with our sparse attention strategy achieves 29 ms inference latency, which is nearly 2 $\times$ faster than the previous bucketing-based approach~\cite{fan2021embracing} while bringing +0.54 performance gain on L2 mAPH (\textit{i.e.}, 60.86 $\rightarrow$ 61.40). That verifies our sparse attention strategy is much more efficient than standard attention in sparse data processing. 

We further ablate the effectiveness of rotated set partitioning in Table~\ref{tab:ab_set_part}. Several baselines are designed for comparison. Random: randomly sample non-empty voxels inside each window. Sparse: voxels of each subset are evenly distributed across the window region. Regional: a non-rotating variant of our partition approach. These methods can be implemented by controlling inner-window voxel ID (See \S \ref{sec:dynaatt}). As shown in $1^{st}$ and $3^{rd}$ rows, the results of our regional partition are +0.83 L\_2 mAPH higher than random sampling due to the better encoding of part-aware geometric information. Additional rotated configuration provides connections across subsets, further boosting the performance of 3D perception.

Table~\ref{tab:analyis_att} and \ref{tab:hybridwin} show the results of different hybrid window shapes. Our method can greatly reduce the padding cost and deliver good performance-efficiency trade-off. Notably, we observe that attention is efficient actually, thus slight padding doesn't significantly slow down the inference. However, frequent function calls to attention will increase the latency due to additional overheads, (\textit{e.g.}, memory access). 

\noindent\textbf{Effect of 3D pooling module.} To demonstrate the effectiveness of our 3D pooling module, we refer several widely-used pooling strategies as baselines and only switch the pooling strategies while all other modules remain unchanged. 
Note that we unify the set size of both two variants 
 (\textit{i.e.}, DSVT-P and DSVT-V) into 36 for a fair comparison, which is different from the best setting reported in Table ~\ref{tab:one_results}.
See appendix for more baseline details and hyper-parameter analyses. 
As shown in Table \ref{tab:ab_pooling}, our pooling operation outperforms all other baselines.  Interestingly, we find that adding key masks on empty voxels will harm the performance of our attention-style pooling operation, which suggests that the empty voxels also encode the object geometry information.

\begin{table}
\begin{center}
    \centering
        \resizebox{0.95\linewidth}{!}{
        \begin{tabular}{cccc}
        \Xhline{0.8pt}\noalign{\smallskip}
        Models &Latency & L2 mAP & L2 mAPH \\
        \noalign{\smallskip}
        \hline
        \noalign{\smallskip}
        Centerpoint-Pillar& 35ms  & 66.0 &62.2\\
        Centerpoint-Voxel & 40ms & 68.2 & 65.8\\
        PV-RCNN++(center)& 113ms & 71.7 & 69.5\\
        Ours(Pillar) & 67ms & 73.2 & 71.0\\
        Ours(Voxel) & 97ms & 74.0 & 72.1\\
        Ours(Pillar+TensorRT) &37ms & 73.2 & 71.0\\
        \Xhline{0.8pt}
        \end{tabular}
        }
\end{center}
\vspace{-16pt}
\caption{The latency and performance on Waymo validation set.}
\label{table:acc_speed}
\vspace{-16pt}
\end{table}

\subsection{Inference Speed} \label{sec:speed}
We present a comparison with other state-of-the-art methods on both inference speed and performance accuracy in Table~\ref{table:acc_speed}. Our pillar variant significantly outperforms PV-RCNN++~\cite{shi2021pv} while achieving a much lower latency, \textit{i.e.}, 67 ms vs 113 ms. After being deployed by NVIDIA TensorRT, our model can achieve a real-time running speed (27 \textit{Hz}), which is almost as fast as the widely-used Centerpoint-Pillar~\cite{yin2021cvpr} with much higher performance (+8.8 on L2 mAPH).  Such running time is enough since a lidar
typically operates at 10 \textit{Hz} to 20 \textit{Hz}. All the experiments are evaluated on the same workstation and environment.

\section{Conclusion}
In this paper, we propose DSVT, a deployment-friendly yet powerful transformer-only backbone for 3D perception. To efficiently handle sparse point clouds, we introduce dynamic sparse window attention, a novel attention strategy that partitions all the sparse voxels into a series of size-equivalent and window-bounded subsets, which can be processed in parallel without any customized CUDA operations. Thus, our proposed DSVT can be accelerated by well-optimized NVIDIA TensorRT, which achieves state-of-the-art performance on various 3D perception benchmarks with real-time running speed. We hope that our DVST can not only be a reliable point cloud processor for 3D perception in real-world applications but also provide a potential solution for efficiently handling sparse data in other tasks. 

\section{Acknowledgement}
This work is supported by National Key R\&D Program of China (2022ZD0114900) and National Science Foundation of China (NSFC62276005).

{\small
\bibliographystyle{ieee_fullname}
\bibliography{egbib}
}

\newpage
\appendix
In the supplementary material, we first elaborate on the proof of non-overlapped set partition in \S\ref{sec:proof}, and then provide more implementation details of the network architecture, training schemes, and ablation baselines in \S\ref{sec:imple_app}. Finally, we present more ablative studies for hyper-parameter analyses in \S\ref{sec:hyana} and visualization of quantitative results in \S\ref{sec:viz}. We also discuss the difference of Axis-attention in \S\ref{sec:axis_att}  and limitation of DSVT in \S\ref{sec:limit}.

\section{Proof of Non-overlap Set Partition} \label{sec:proof}
\noindent \textbf{Definition A.1} (\textbf{\textit{Dynamic set partition}}) $N$ is the number of non-empty voxels for a specific window, and $\tau$ is the maximum
number of sparse voxels allocated to each local set. The required number of sub-sets in this
window is computed as follows,
\begin{equation}
    S = \lfloor \frac{N}{\tau} \rfloor + \mathbb{I}[(N~\%~\tau) > 0].
\end{equation}
For the $j$-th set (denoted as $\mathcal{Q}_j = \{q^j_k\}^{\tau-1}_{k=0}$), the index of voxel can be computed:
\begin{equation}
  q^j_k = \left\lfloor  \widetilde{q}^j_k  \right\rfloor = \left\lfloor\frac{ (j \times \tau + k)}{S \times \tau} \times N \right\rfloor,  ~~\text{for}~~ k = 0, ..., \tau - 1.
  \label{eq:index_calculate}
\end{equation}
The following theorems show that our algorithm formulation in the main paper satisfies all the necessary concepts of the proposed dynamic set partition. The case of $N=S \times \tau$ is trivial. Now suppose the case of $(S-1) \times \tau < N < S \times \tau$. 

\noindent \textbf{Theorem A.2.} (\textit{\textbf{Non-overlap}}): \textit{For any two local sets, $0 \leq i, j \leq S-1$, then $\mathcal{Q}_{i} \cap \mathcal{Q}_{j} = \emptyset$.}

\noindent $Proof.$ Obviously, this theorem can be converted to verify the non-overlap of two neighboring sets. Specifically, for any $1 \leq j \leq S-1$, $\mathcal{Q}_{j-1} \cap \mathcal{Q}_{j} = \emptyset$. We formulate $q^j_0$ (the first voxel index in ${Q}_{j}$) as follows,
\begin{equation}
  q^j_0 = \left\lfloor\frac{j \times N}{S} \right\rfloor = \left\lfloor{N^j_0 + \frac{k^j_0}{S}}\right\rfloor = N^j_0,  ~~0 \leq k^j_0 \leq S - 1.
\label{eq:3}
\end{equation}

\noindent Then we can compute the last voxel index in ${Q}_{j-1}$:
\begin{equation}
\begin{aligned}
  q^{j-1}_{\tau-1} & = \left\lfloor\frac{(j-1) \times \tau + \tau - 1}{S \times \tau} \times N \right\rfloor \\ 
                   & = \left\lfloor \frac{j \times N}{S} - \frac{N}{S\times \tau} \right\rfloor = \left\lfloor N^j_0 + \frac{k^j_0}{S} - \frac{N}{S\times \tau}\right\rfloor.
\end{aligned}
\end{equation}

\noindent Note that $(S-1) \times \tau < N < S \times \tau$, thus $\frac{S-1}{S} < \frac{N}{S\times \tau} < 1$ and $0 \leq \frac{k^j_0}{S} \leq \frac{S-1}{S}$. So we have
\begin{equation}
\begin{aligned}
  \left\lfloor N^j_0 - 1\right\rfloor  < \left\lfloor N^j_0 + \frac{k^j_0}{S} - \frac{N}{S\times \tau}\right\rfloor  < \left\lfloor N^j_0 \right\rfloor. 
\end{aligned}
\end{equation}

\noindent To this end, we can prove that $q^{j-1}_{\tau-1} = N^j_0 - 1$. For any $1 \leq j \leq S-1$, $q^{j-1}_{\tau-1} \neq q^{j}_{0}$, thus $\mathcal{Q}_{j-1} \cap \mathcal{Q}_{j} = \emptyset$.

\noindent \textbf{Theorem A.3.} (\textit{\textbf{Completeness}}): \textit{$\mathcal{Q}_{0} \cup \mathcal{Q}_{1} \cup ...\cup \mathcal{Q}_{S-1} = U$, where $U=\{0, 1, ..., N-1\}$.}

\noindent $Proof.$ If $u\in \{0, 1, ..., N-1\}$ but $u\notin U$, there must exist two continuous indexes $q^{j}_{k-1}$ and $q^{j}_{k}$ satisfying $q^{j}_{k-1} < u $ and $q^{j}_{k} \geq u+1$. Thus we have:
\begin{equation}
\begin{aligned}
\frac{(j \times \tau + k)}{S \times \tau} \times N - \frac{(j \times \tau + k-1)}{S \times \tau} \times N > 1.
\end{aligned}
\end{equation}
This yields a contradiction, because $N < S \times \tau$, and concludes the proof.

\noindent \textbf{Theorem A.4.} (\textit{\textbf{Equivalent}}): \textit{For any j-th subset, $0 \leq j \leq S-1$, we have $\lfloor \frac{N}{S} \rfloor \leq |\mathcal{Q}_{j}| \leq \lfloor \frac{N}{S} \rfloor + 1$. $|\mathcal{Q}_{j}|$ denotes the number of valid and unique voxels belonging to j-th set, which needs to be distinguished from $\tau$.}

\noindent $Proof.$ We reformulate $N$ as:
\begin{equation}
\begin{aligned}
  N  &= \left\lfloor\frac{N}{S} \right\rfloor \times S + res  = l \times S + res,
\end{aligned}
\end{equation}
where $l=\left\lfloor\frac{N}{S} \right\rfloor $ and $0\leq res \leq S-1$. In the case of $res=0$, we have:
\begin{equation}
\begin{aligned}
   q^{0}_{0}=0, ~q^{1}_{0}=l,~..., ~q^{S-1}_{0}=(S-1)\times l.
\end{aligned}
\end{equation}
Following the proof of Theorem A.2, we can have:
\begin{equation}
\begin{aligned}
   q^{0}_{\tau-1} &=l-1, ~q^{1}_{\tau-1}=2\times l -1,~..., ~q^{S-1}_{\tau-1}&=S\times l - 1,
\end{aligned}
\end{equation}
which indicates for any $j$, we obtain $|\mathcal{Q}_{j}|=l=\lfloor \frac{N}{S} \rfloor$. In the case of $res\neq0$, we compute the difference between   $ \widetilde{q}^{j}_{0}$ and $ \widetilde{q}^{j+1}_{0}$:
\begin{equation}
\begin{aligned}
   \widetilde{q}^{j+1}_{0} -  \widetilde{q}^{j}_{0} & = \frac{(j+1) \times \tau}{S \times \tau} \times N - \frac{j \times \tau}{S \times \tau} \times N  \\
                          & = \frac{N}{S} = l + \frac{res}{S}.
\end{aligned}
\label{eq:res}
\end{equation}
By Eq.~\eqref{eq:3} and Eq.~\eqref{eq:res}, we have
\begin{equation}
\begin{aligned}
  q^{j+1}_0 & = \left\lfloor\frac{(j + 1) \times N}{S} \right\rfloor =  \left\lfloor\frac{j \times N}{S} + \frac{N}{S}\right\rfloor \\
            & = \left\lfloor{N^j_0 + \frac{k^j_0}{S} + l + \frac{res}{S}}\right\rfloor = N^j_0 + l +  \left\lfloor\frac{res+k^j_0}{S} \right\rfloor.
\end{aligned}
\end{equation}
Note that $0 \leq k^j_0 \leq S - 1$ and $0 < res \leq S - 1$, therefore we can easily obtain:
\begin{equation}
\begin{aligned}
 N^j_0 + l \leq q^{j+1}_0 \leq N^j_0 + l + 1.
\end{aligned}
\end{equation}
Finally, following the proof of Theorem A.2, we have:
\begin{equation}
\begin{aligned}
 N^j_0 + l - 1 \leq q^{j}_{\tau-1} \leq N^j_0 + l ,
\end{aligned}
\end{equation}
which implies for any $j$, we have $l \leq |\mathcal{Q}_{j}| \leq l+1$ and concludes the proof.

\section{Implementation Details} \label{sec:imple_app} 
In this section, we provide more implementation details about network architecture (\S\ref{subsec:imple}), ablation baselines (\S\ref{subsec:ab_baseline}) and training schemes (\S\ref{subsec:schemes}). 
\subsection{Network Architecture.} \label{subsec:imple} 
\subsubsection{3D Perception on Waymo}
As mentioned in the main paper, our detection approach follows the framework of CenterPoint-Pillar~\cite{yin2021cvpr} and only appends our DSVT before BEV backbone while other components remained unchanged. 

\noindent \textbf{DSVT-P} is a single-stride pillar-based sparse backbone, which adopts the pillar size of (0.32m,
0.32m, 6m) with four DSVT blocks. Each block is equipped with two DSVT layers with different set partitioning configurations, (\emph{i.e}, X-Axis Partition and Y-Axis Partition). The DSVT layers contains a rotated set based Multi-Head Self-Attention (MHSA) module, followed by a 2-layer MLP with GELU nonlinearity in between. A LayerNorm (LN) layer is applied after each MHSA module and each MLP, and a residual connection is applied after each module. All the attention modules are equipped with 8 heads, 192 input channels, and 384 hidden channels. The hybrid window sizes are set to (12, 12, 1) and (24, 24, 1) by default, and the maximum number of voxels belonging to each set ($\tau$) is 36, as introduced in main paper.

\noindent \textbf{DSVT-V} is a voxel-based variant of our proposed backbone, which follows the pillar-based framework and splits along the Z-Axis. The input voxel size is (0.32m, 0.32m, 0.1875m). Moreover, its backbone also has four stages with block numbers \{1, 1, 1, 1\} and the number of voxels along the Z-Axis is reduced by our attention-style 3D pooling module with the stride \{4, 4, 2\}. The window sizes along the Z-Axis are \{32, 8, 2, 1\}, which covers all of the Z-Axis. Different from DSVT-P, to adapt more voxels, $\tau$ is set to 48.
\subsubsection{3D Perception on nuScenes} 
3D object detection and BEV Map Segmentation both utilizes DSVT-P in nuScenes benchmark. We set window size and set the maximum number of tokens assigned to each set ($\tau$) to (30, 30, 1) and 90, respectively. The attention modules in use were equipped with 8 heads, 128 input channels, and 256 hidden channels. 
\subsection{Ablation Baselines.} \label{subsec:ab_baseline}
\subsubsection{ResBackbone1x}
ResBackbone1x is built upon sparse convolution (Spconv 2.0)~\cite{spconv2022}, a widely used auto-differentiation library for sparse tensors. This baseline adopts the same network designs (\emph{i.e.}, depth, width, and kernel size) as VoxelResBackBone8x implemented by OpenPCDet~\cite{openpcdet2020} except for replacing all the downsampling SparseConv blocks with conventional SubMConv to hold the single stride architecture. The input voxel size is set to (0.32m, 0.32m, 6m), which is the same as our DSVT pillar version.
For a fair comparison, this variant only substitutes the DSVT sparse backbone with ResBackbone1x while other settings remained unchanged, (\emph{e.g.}, detection head, loss functions, and post-processing). As shown in Table \ref{table:backbone_compare}, thanks to the single stride design, this baseline is very strong, which is +4.89 better than the original CenterPoint-Voxel(8x)~\cite{yin2021cvpr} and +2.80 higher than its residual modification version on L2 mAPH. Even on such a strong baseline, our DSVT performs +1.78 better, which demonstrates its powerful modeling ability.
\begin{table}
\begin{center}
    \centering
        \begin{tabular}{c|ccc}
        \Xhline{0.8pt}\noalign{\smallskip}
        \multirow{2}{*}{Backbone}& \multirow{2}{*}{\#param.}& \multicolumn{2}{c}{LEVEL\_2 (3D)} \\
        &  & mAP&mAPH\\
        \noalign{\smallskip}
        \hline
        \noalign{\smallskip}
        VoxelBackBone8x\dag& 58M&64.51&61.92\\
        VoxelResBackBone8x\dag& 80M& 66.47&64.01\\
        ResBackbone1x & 88M & 69.61 &66.81\\
        DSVT(Pillar, dim128) & 71M  &\textbf{71.14} &\textbf{68.59}\\
        \Xhline{0.8pt}
        \end{tabular}
\end{center}
\vspace{-10pt}
\caption{Comparison with sparse convolution.~ $\dag$ denotes the results implemented by OpenPCDet~\cite{openpcdet2020}. All models are trained on 20\% Waymo data with 30 epochs.}
\label{table:backbone_compare}
\vspace{-10pt}
\end{table}

\subsubsection{3D Pooling}
\noindent \textbf{Linear.} As for a specific downsampling sparse region, we first convert it into dense and flatten it to a vector with fixed length. Then a one-layer MLP is applied to project it. Finally, a layer normalization is adopted after MLP module.

\noindent \textbf{Max Pooling.} Similar to the linear variant, after being converted into a dense format, the native max-pooling operation is applied on voxel dimension for processing downsampling.

\noindent \textbf{Attention+Mask.} This variant follows the same design as our attention-style 3D pooling module except for adding key padding masks of the empty space in the pooling region.
\subsection{Training and Inference Schemes.}\label{subsec:schemes}
\subsubsection{Waymo}
\noindent \textbf{One Stage Detection.} As mentioned in the main paper, we follow the same training schemes as \cite{yin2021cvpr} to optimize
the model using Adam~\cite{kingma2014adam} optimizer with weight decay 0.05, one-cycle learning rate policy~\cite{onecyc}, and max learning rate 3e-3. All the models are trained with batch size 24 for 24 epochs on 8 NVIDIA A100 GPUs. During inference, following~\cite{YihanHu2022AFDetV2RT,shi2022pillarnet}, we use class-specific NMS with the IoU threshold of 0.7, 0.6 and 0.55 for vehicle, pedestrian and cyclist, respectively. Besides, we also use the ground-truth copy-paste data augmentation during training and disable this data augmentation in the last one epoch following~\cite{fan2022fully} (\emph{e.g.}, using the fade strategy).

\noindent \textbf{Two Stage Detection.} The two-stage of our DSVT is built upon CT3D~\cite{sheng2021improving} and trained separately. We fix the 1$^{st}$-stage model and finetune the 2$^{nd}$-stage refinement module for 12 epochs with the same training schedule. 
\subsubsection{NuScenes}
\noindent \textbf{3D Object Detection.} We follow the same training scheme adopted in Transfusion-L~\cite{bai2022transfusion}. All the models are trained by AdamW optimizer with weight decay 0.05, one-cycle learning rate policy, max learning rate 5e-3, and batch size 32 for 20 epochs.We adopt the same fade strategy in \cite{bai2022transfusion} in last 5 epochs.

\noindent \textbf{BEV Map Segmentation.} We also adopt the same training strategy as BEVFusion~\cite{liu2022bevfusion}, including training epoch, learning rate and hyper-parameter of optimizer.

\section{Hyper-parameter Analyses} \label{sec:hyana} 
Our DSVT also works well in a wide range of hyper-parameters, such as the set size and network depth. All the experiments are trained on 20\% Waymo training data with 30 epochs.

\noindent \textbf{Set Size.}
Table~\ref{tab:setsize} shows the performance of our approach with different set sizes. With the increase of the set size (from 24 to 36 in DSVT-P, 36 to 48 in DSVT-V), the performance gradually improves. However, a very large set size will also slightly decrease the mAP/mAPH. We argue that our regional local set attention can better encode the part-aware geometric information, which enhances the performance of tiny objects. The large set coverage may involve lots of noise points. Thus, we set the set sizes to 36 and 48 for our DSVT-P and DSVT-V respectively.

\begin{table}
\begin{center}
    \centering
        \begin{tabular}{ccccc}
        \Xhline{0.8pt}\noalign{\smallskip}
        \multirow{2}{*}{DSVT-P}& \multirow{2}{*}{DSVT-V} &\multirow{2}{*}{Size} & \multicolumn{2}{c}{LEVEL\_2 (3D)} \\
        &  & & mAP&mAPH\\
        \noalign{\smallskip}
        \hline
        \noalign{\smallskip}
        $\checkmark$& &24& 70.71 & 68.14\\
        $\checkmark$& &36& \textbf{71.14} & \textbf{68.59}\\
        $\checkmark$& &48& 70.95 & 68.43\\
        \toprule
        &$\checkmark$&36& 71.65 & 69.31\\
        &$\checkmark$&48& \textbf{72.01} & \textbf{69.67}\\
        &$\checkmark$&60& 71.90& 69.60\\
        \Xhline{0.8pt}
        \end{tabular}
\end{center}
\vspace{-15pt}
\caption{Effect of set size.}
\label{tab:setsize}
\vspace{-10pt}
\end{table}

\noindent \textbf{Network Depth.}
DSVT is relatively shallow by design thanks to the large receptive fields of the transformer architecture. As shown in Table \ref{tab:netdep}, we provide 
the results with a greater number of DSVT blocks for investigating the influence of network depth. We observe that the performance is gradually saturated with the increase of block number. A deeper network will decrease the running speed. Considering the trade-off between the computation cost and performance improvement, we choose 4 blocks as the default setting. 

\begin{table}
\begin{center}
    \centering
        \begin{tabular}{ccc}
        \Xhline{0.8pt}\noalign{\smallskip}
        \multirow{2}{*}{\# of Blocks}& \multicolumn{2}{c}{LEVEL\_2 (3D)} \\
         & mAP&mAPH\\
        \noalign{\smallskip}
        \hline
        \noalign{\smallskip}
        2& 70.66 & 68.10\\
        4&71.14 & 68.59\\
        6&71.12 & 68.56\\
        8& \textbf{71.24} & \textbf{68.68}\\
        \Xhline{0.8pt}
        \end{tabular}
\end{center}
\vspace{-15pt}
\caption{Effect of network depth.}
\label{tab:netdep}
\vspace{-10pt}
\end{table}

\begin{figure*}
  \centering
  \includegraphics[width=0.95\linewidth]{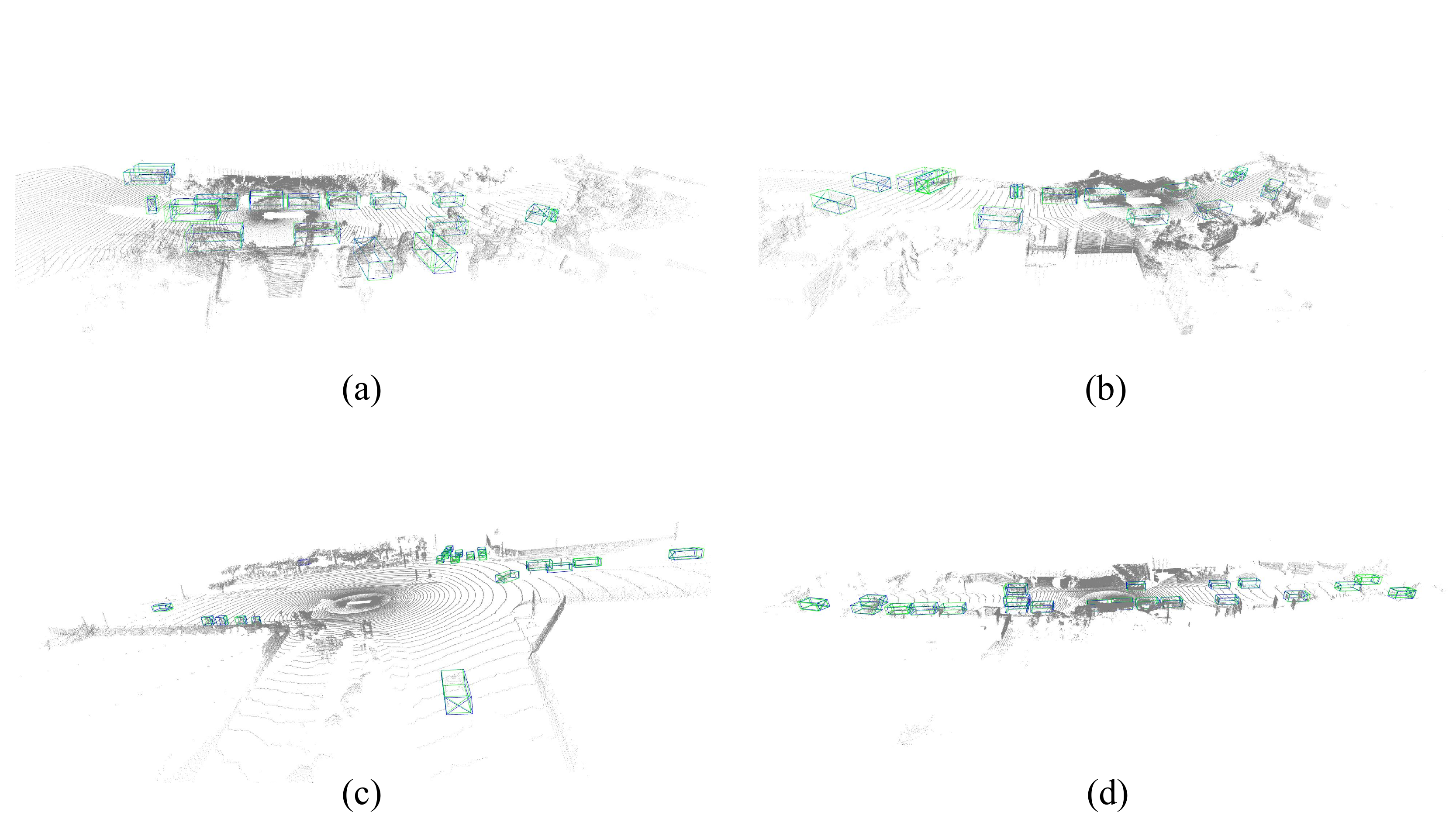}
  \vspace{-6pt}
    \caption{Qualitative visualization on Waymo validation set. Blue boxes and green boxes are ground-truth and predictions, respectively.}
  \label{fig:visual}
  \vspace{-14pt}
\end{figure*}

\section{Qualitative Results} \label{sec:viz}
We visualize the qualitative results on  Waymo Validation Set in Figure \ref{fig:visual}. Thanks to the large receptive field of Transformer and fine-grained geometric information provided by the attention-style 3D pooling module, our DSVT performs well on the large scenes and can locate 3D objects with sparse points accurately.

\section{Compared to Axial-attention} \label{sec:axis_att}
Our method cannot be considered as an extension of Axial-attention~\cite{wang2020axial}. DSVT is specifically designed for efficiently processing sparse data in parallel with dynamically assigned and size-equivalent local sets. The axis-based rotated partitioning is a replaceable strategy for intra-window fusion (please see Table \ref{tab:ab_set_part} in the main paper for alternative strategies). In contrast, axial-attention~\cite{wang2020axial,bertasius2021space} aims to reduce the attention computation cost due to the dense data (\emph{e.g.}, 2D image, or video) and enlarge receptive field by axis-based factorization.

\section{Limitation} \label{sec:limit}
Although our method achieves promising performance and running speed on Waymo Open dataset, there are still some limitations. DSVT mainly focuses on point cloud processing for 3D object detection in outdoor autonomous driving scenarios, where objects (\emph{i.e.}, car, pedestrian and cyclist) are distributed on a 2D ground plane. It is still an open problem to design more general-purposed backbones for the 3D community.

\end{document}